# Mixed Graphical Models for Causal Analysis of Multi-modal Variables

**Authors:** Andrew J Sedgewick[1,2], Joseph D. Ramsey[4], Peter Spirtes[4], Clark Glymour[4], Panayiotis V. Benos[2,3,*]

**Affiliations:**

[1]Department of Computational and Systems Biology, University of Pittsburgh School of Medicine, Pittsburgh, Pennsylvania, USA.

[2]Joint Carnegie Mellon University-University of Pittsburgh PhD Program in Computational Biology, Pittsburgh, Pennsylvania, USA.

[3]Department of Bioengineering, University of Pittsburgh, Pittsburgh, Pennsylvania, USA.

[4]Department of Philosophy, Carnegie Mellon University, Pittsburgh, Pennsylvania, USA.

*To whom correspondence should be addressed: benos@pitt.edu

## Abstract

Graphical causal models are an important tool for knowledge discovery because they can represent both the causal relations between variables and the multivariate probability distributions over the data. Once learned, causal graphs can be used for classification, feature selection and hypothesis generation, while revealing the underlying causal network structure and thus allowing for arbitrary likelihood queries over the data. However, current algorithms for learning sparse directed graphs are generally designed to handle only one type of data (continuous-only or discrete-only), which limits their applicability to a large class of multi-modal biological datasets that include mixed type variables. To address this issue, we developed new methods that modify and combine existing methods for finding undirected graphs with methods for finding directed graphs. These hybrid methods are not only faster, but also perform better than the directed graph estimation methods alone for a variety of parameter settings and data set sizes. Here, we describe a new conditional independence test for learning directed graphs over mixed data types and we compare performances of different graph learning strategies on synthetic data.

# INTRODUCTION

**Background.** In the era of Big Data, many datasets routinely collected, including commonly studied biological and biomedical data, are multi-modal: that is, they include both discrete and continuous variables. The sizes of relevant databases containing these data have become enormous. This demands the aid of fast, accurate, computerized search methods for predicting causal relations. Directed probabilistic graphical models (**PGM**s) can represent these causal relationships based on the conditional (in)dependencies of the data. In addition, these models fit a joint probability distribution to high-dimensional observations. Causal graphs are represented as directed graphs or collections of directed graphs with identical conditional independencies. The resulting graph can provide guidance for future intervention experiments and are useful for classification and prediction of outcomes of certain target variables of interest. A number of methods for learning directed graphs have been developed in the past, but they typically assume (for proof of asymptotic correctness) that all variables are of the same distribution type: categorical (multinomial), Gaussian, conditional Gaussian, or linear non-Gaussian.

Several groups have developed methods to learn undirected graphs over mixed data types (Tur and Castelo 2012; Cheng et al. 2013a; Fellinghauer et al. 2013; Lee and Hastie 2013; Chen et al. 2014; Yang et al. 2014); and directed graphs over mixed variables under certain distributional assumptions (Bøttcher 2001; Romero et al. 2006). One of the popular methods for learning undirected mixed graphical models (**MGM**) is Lee and Hastie (Lee and Hastie 2013). We recently presented several improvements of the Lee and Hastie method (Sedgewick et al. 2016). A major problem of the undirected (i.e., non-causal) graphs, apart from the lack of direction of the represented interactions, is that they are "moralized" graphs; meaning, the parents of a variable are themselves always connected. This can create a large number of false positive edges.

In this paper, we present and test new methods for learning a directed MGM. The main idea of our approach is to first learn the undirected graph and then prune-and-orient this graph using methods derived from existing algorithms for directed graph learning. We use our modified version of Lee and Hastie to learn the undirected graph. For the prune-and-orient step we use the strategies implemented in PC-stable and CPC-stable (Colombo and Maathuis 2014). PC-stable is a modification of PC (Spirtes and Glymour 1991), the oldest correct algorithm for searching for directed acyclic graphs when there are no feedback relations and no unrecorded common causes and sampling is independent and identically distributed.

**Related work.** Recently, learning a sparse undirected graph structure over multi-modal datasets has attracted attention (Bøttcher 2001; Romero et al. 2006; Tur and Castelo 2011; Cheng et al. 2013b; Fellinghauer et al. 2013; Lee and Hastie 2013; Chen et al. 2014; Yang et al. 2014). There is publically available software for several of these methods (Tur and Castelo 2011; Fellinghauer et al. 2013; Lee and Hastie 2013). The Tur and Castelo method is not able to learn connections between categorical variables. This approach is appropriate for the study of expression quantitative trait loci (eQTLs), but does not allow for analysis of downstream discrete clinical variables, for example. A number of proposals suggest a nodewise regression approach for learning networks over a variety of distributions of continuous and discrete variables (Cheng et al. 2013b; Fellinghauer et al. 2013; Chen et al. 2014), Lee and Hastie (Lee and Hastie 2013) propose optimizing the pseudolikelihood of a mixed distribution over Gaussian and categorical variables. We developed our algorithms using Lee and Hastie's method as a starting point, both because we will only look at Gaussian and categorical variables in this study and because their approach involves learning fewer parameters than with nodewise regression methods.

The idea of using an undirected method to estimate a superstructure of the true graph, and then restricting the search space of a directed search algorithm to the superstructure has previously been studied for continuous, possibly non-Gaussian data with linear interactions between nodes (Loh and Bühlmann 2014). Like our proposed method, Loh and Bühlmann first find a undirected graph which serves as an estimate of the moralization of the true graph, and then use this undirected graph as an estimate for a directed search method. The two primary differences between this study and our proposals are that Loh and Bühlmann only look at continuous data in their study, and that the directed search is a score-based method while we focus on constraint-based directed search methods here.

Adapting score-based methods to mixed data is a challenging problem that we are very interested in. The concept behind score-based methods is to efficiently search over the space of DAGs to find the structure that has the best score given the data. In general, these scores take advantage of the fact that joint probability distributions represented by DAGs are factorizable, so adding or subtracting edges from the estimated graph only require re-calculating scores of the incident nodes. Scores are usually related to likelihood calculations, for example, the Bayesian information criterion (BIC) is commonly used for continuous data and is calculated by penalizing the log-likelihood for the degrees of freedom and sample size. The challenge is to find a mixed score that is factorizable and efficient to compute. This is an open area of research, but should a useable score for mixed data be developed, these methods can take advantage of an initial undirected graph by using it to restrict the search space of the score-based search algorithm.

## MATERIALS AND METHODS

### Simulated data

We simulated data from low- and high-dimensional networks with 50 different directed graph topologies each, randomly selected using TETRAD (version 5.3.0, https://github.com/cmu-phil/tetrad), a Java package for causal modeling that uses linear or non-linear structural equation models (SEMs) to generate data from network distributions. The low dimensional datasets consisted of 500 samples each, drawn from network structures of 50 variables: 25 Gaussian and 25 3-level categorical. The high dimensional datasets consisted of 100 samples drawn from network structures of 200 variables: 100 Gaussian and 100 3-level categorical. The structures are sampled uniformly from the space of all directed acyclic graphs (DAGs) with maximum node degree of 10 and a maximum of average node degree of 2. The low dimensional dataset allows to test the efficiency of the algorithms when the study is well powered; while the high dimensional dataset test the efficiency of the algorithms when the number of samples is small compared to the number of variables, a condition that is frequently present in many biomedical applications.

The relationships between variables in all datasets are set up in a similar fashion to Lee and Hastie (Lee and Hastie 2013). Here, for an edge $X \rightarrow Y$ we refer to X as the parent and Y as the child. Parents of the Gaussian variables contribute linearly to the mean of each child; the value of continuous parents is multiplied by an edge parameter and the value of discrete parents is associated with an edge parameter where a separate edge parameter is specified for each category of the discrete variable. Parents of discrete variables contribute log-linearly to the probabilities of each category, with separate parameters for each category of the child variable. With this set up, each edge connecting two continuous variables (*cc*) depends on 1 edge parameter, each edge connecting a continuous and a discrete variable (*cd*) depends on a vector of 3 parameters and edges connecting two discrete variables (*dd*) depend on a 3 by 3 matrix of 9 edge parameters. In order to ensure identifiability, the *cd* parameter vector, and the rows of the *dd* parameter matrix are constrained to sum to 0 leaving these edges with 2 and 6 degrees of freedom, respectively. Edge weights were drawn uniformly from the union of the regions [-1.5, -1] and [1, 1.5]. For *cc* edges the parameter is equal to the weight; for *cd* edge parameters we draw a vector three values uniformly from [0,1] and shift and scale the values so they sum to zero and the largest parameter is equal to the edge weight; for *dd* edge parameters we draw one vector of three values as with *cd* edges and set the rows of the matrix as the three permutations of this vector.

 In the continuous case, zero-mean, Gaussian error terms with standard deviation uniformly drawn from the interval [1, 2], are drawn for every variable and then the variable means are resolved. In DAGs this resolution is trivial as we can start from root nodes with no parents and propagate downwards. To make this process accommodate categorical distributions, we use a uniform draw over [0, 1] as an error term for each discrete variable and this term is used to determine the value of the variable given the probabilities of each category. In generating simulated models, these probabilities that are then updated in the same way as are the means of the continuous variables. This approach ensures convergence of each discrete variable for each sample.

**Conditional independence tests for scoring mixed type edges**

One of the important components of constrained based methods for learning a graph is the edge scoring. This is typically achieved with a hypothesis test for conditional dependence of two variables, $X$ and $Y$, given a conditioning set of variables, $S$. The null hypothesis is that $X$ and $Y$ are independent given $S$, which is denoted by $X \perp Y \mid S$. By definition, if this null hypothesis is true:

$$P(X, Y|S) = P(X|S)P(Y|S)$$

Rearranging, we find:

$$P(X|S) = \frac{P(X, Y|S)}{P(Y|S)} = P(X|Y, S)$$

So, in order to test $X \perp Y \mid S$ it suffices to test if $P(X|S) = P(X|Y, S)$ which is done via likelihood ratio test (LRT) of two regressions. This test is known to follow the chi-squared distribution.

$$2 \ln \left( \frac{L(\theta_{XYS})}{L(\theta_{XS})} \right) \sim \chi^2(d_X d_Y)$$

Where $\theta$ represents the regression coefficients to model $X$ given $S$ with and without $Y$ as an additional independent variable. This test is used by PC-stable (Colombo and Maathuis 2014) but we modify it to accommodate mixed data types. Specifically, we define the degrees of freedom, $d_X$ and $d_Y$, of each variable to be (a) 1 if the variable is continuous and (b) the number of categories minus 1 if the variable is categorical. Although this description uses regressions with $X$ as the dependent variable, the same reasoning allows us to use $Y$ as the dependent variable instead.

The regressions in this test allow us to formulate this test so that any of the variables can be continuous or categorical. We preform linear or multinomial logistic regressions if the dependent variable is continuous or categorical, respectively. Because of this, if $X$ and $Y$ are of different variable types, we have a choice of whether $X$ or $Y$ should be the independent variable that determines whether we perform logistic or linear regressions. Our own experiments (see Supplementary Material) and observations in previous studies (Chen et al. 2014) suggest that a linear regression will give a more accurate test result than a logistic regression for these continuous-discrete edges. To handle any dependent categorical variables in the regression, use the standard practice of converting each $k$-level categorical variable to $k$-1 binary variables.

It is also possible to conduct these tests by regressing $Y$ and $S$ onto $X$, and using a $t$-test to determine if the regression coefficient of $Y$ is significantly different from 0. In the continuous setting, the t-test and LRT give virtually identical results, but not with discrete or mixed data. If $Y$ is categorical this procedure requires performing a test on each dummy variable associated with $Y$ and then combining them using Fisher's method. The main advantage of using $t$-tests over the LRT that it only requires one regression instead of two, so it is significantly faster. The downside is that in our experiments we found that it had less power to detect true edges (data not shown), and was less robust at low sample sizes, particularly on edges that required a logistic regression. Because of this we will work exclusively with the LRT based test here.

**Graph search methods**

Given an edge scoring method, graph search algorithms are efficient heuristics to search the exponential space of all possible graph configurations. Here we test two popular algorithms, PC-stable and CPC-stable (Colombo and Maathuis 2014), which are derived from the PC algorithm(Spirtes and Glymour 1991).

The **PC** algorithm and its descendants depend on conditional independence decisions that are made by a user-specified test and the $\alpha$ threshold (described below). PC starts with a complete graph and in step 1 it sequentially tests all edges for independence given conditioning sets of increasing size. Starting with the empty set, these conditioning sets are subsequently made up of every set (of the given size) of common neighbors of the two nodes incident to the edge being tested. Edges that are found to be conditionally independent are immediately removed and not considered in future tests. When an edge is removed, the conditioning set that lead to the independence decision is saved. Step 2 directs edges based on the fact that common neighbors of nodes incident to a removed edge that are not in the conditioning set must be in a v-structure ($X \rightarrow Z \leftarrow Y$). It is possible that two implied v-structures will induce conflicting edge directions. Step 3 further directs edges based on a set of rules that ensure the directions will not induce any cycles or new v-structures (Spirtes et al. 2000). **PC-stable** modifies PC by waiting to update the edge removals in phase 1 until all tests for a given conditioning set size are completed. This leads to an output that is independent of variable ordering and allows for parallelization of the independence tests. Also, in the TETRAD implementation of PC-stable direction conflicts result in a bi-directed edge: $X \leftrightarrow Y$.

**CPC-stable** (Colombo and Maathuis 2014) is the variable order independent variant of **Conservative PC** (Ramsey et al. 2006) which revises step 2 of PC, described above to perform conditional independence tests with all possible conditioning sets between two nodes, $X$ and $Y$, that have had an edge between them removed. The conditioning sets are determined by taking subsets of neighbors of the two nodes found in the skeleton graph returned by step 1 of PC. For any node, $Z$ that is incident to both $X$ and $Y$, the v-structure is only predicted if $Z$ is not in any separating set $S$ such that $X \perp Y \mid S$. Otherwise no direction is predicted from this triplet of nodes. If $Z$ participates in some sets that result in the conditional independence of $X$ and $Y$ and some that result in a conditional dependence, the ambiguity is recorded. Since the change to the PC algorithm takes place after adjacency has been determined, PC and CPC algorithms will produce the same adjacency predictions.

In this paper we test PC-stable and CPC-stable using the modified LLR test for mixed data types we describe above. In addition, instead of starting from a fully connected graph, we present a 2-step approach, where we first calculate an undirected graph as in Sedgewick *et al* (Sedgewick et al. 2016) and use PC-stable and CPC-stable with the undirected MGM graph as starting point. We call these algorithm variants **MGM-PCS** and **MGM-CPCS**, respectively.

**Stability Selection**

We tested **CPSS** (Shah and Samworth 2013), which is a variation of the Stability Selection (Meinshausen and Bühlmann 2010) that both loosens the assumptions on the selection procedure (i.e. our network prediction algorithms are "selecting" edges), and tightens the bounds on the error rate, allowing for a less stringent threshold. Besides the obvious benefit of tighter bounds, the loose assumptions are especially attractive to us, as we would like to be able to substitute a variety of algorithms without worrying about violating the theoretical framework of the method. This method works by learning networks over subsamples of the data and counting how many times an edge appears, which is similar to the StEPS approach we developed for learning the undirected MGM graphs (Sedgewick et al. 2016). Rather than calculating network instabilities from these empirical edge probabilities, edges are selected by simply thresholding the probabilities. The threshold is calculated from the number of subsamples, the average number of selected edges, and the number of variables using Shah and Samworth's procedure. The user specifies an error control rate where errors are defined as edges that have a lower than random probability of being selected in a given subsample. We ran CPSS in conjunction with MGM-PCS and MGM-CPCS with $\alpha = .05$ and $\lambda = .1$ for the LD dataset, and with $\lambda = .2$ for the HD dataset with error rates $q \in \{.001, .01, .05, .1\}$.

**Edge recovery evaluation**

To evaluate network estimation performance, we compare the Markov equivalence classes of the estimated and true networks. Markov equivalence classes represent the variable independence and conditional relationships for an acyclic directed graph by removing the direction from edges that are free to point in either direction without altering the independence relationships in the network. For example, directed graphs $X \rightarrow Y \rightarrow Z$, and $X \leftarrow Y \leftarrow Z$ both have the Markov equivalence class $X - Y - Z$ while the graph $X \rightarrow Y \leftarrow Z$ (v-structure) would remain the same when converted to a Markov equivalence class. Thus, Markov equivalent graphs share the same variables, have the same adjacencies, and imply the same independence and conditional independence relations among their variables. We also consider performance on skeleton estimation, (i.e. the set of node adjacencies, without edge orientations).

We use standard classification statistics to evaluate the recovery of the undirected adjacencies from the skeleton of the true graph. Precision, also known as true discovery rate or positive predictive value is the proportion of predicted edges that are found in the true graph. Recall, also known as sensitivity or true positive rate, is the proportion of edges in the true graph that were found in the predicted graph. For direction recovery we use these same statistics applied to the recovery of only the directed edges in the Markov equivalence class of the true graph. So, in the context of direction recovery, precision is the number of directed edges in the predicted graph that are found in the true graph out of the total number of directed edges in the predicted graph. Bi-directed edges are treated as undirected edges for these statistics because they do not give an indication of which edge direction is more likely.

We use the Matthews correlation coefficient (MCC) (Matthews 1975) as a measure for overall recovery performance that strikes a balance between precision and recall. The MCC is a formulation of Pearson's product-moment correlation for two binary variables (i.e. true edge indicators and predicted edge indicators). In addition, we use the structural Hamming distance (SHD) (Tsamardinos et al. 2006) as a combined measure of adjacency and direction recovery. The SHD is the minimum number of edge insertions, deletions, and directions changes, where only undirected edges are inserted or deleted, to get from the true Markov equivalence class to the estimated equivalence class.

## RESULTS AND DISCUSSION

**Simulation experiments.** In order to determine which algorithms have the most general applicability, we performed experiments using two different dataset sizes and randomly drawn DAG structures. In addition, since optimal parameter setting is a difficult problem that may depend on the needs and goals of the user, we studied a range of possible parameter settings to show the relationship between these settings and edge recovery performance.

*Adjacency Recovery.* **Figure 1** shows the (undirected) adjacency recovery performance of PC-stable, MGM-PCS and CPSS on the HD dataset. CPC-stable and MGM-CPCS are not shown because they have the same adjacency predictions as the PC algorithms. Settings of $\lambda < .2$ for the MGM-PCS algorithm are omitted from the figure because they mostly overlap with the PC-stable curves. Despite the apparent overlap, these denser MGM structures do cause a slight decrease in the precision of MGM-PCS compared to PC-stable, although this difference is not significant at any of the tested settings. For example, at $\alpha = .05$ and $\lambda = .14$, MGM-PCS has an average precision of .739 (standard error of .0057) compared to PC-stable which achieves mean precision of .744 (standard error is .0055). We expect that in the limit of $\lambda \rightarrow 0$, MGM-PCS becomes equivalent to PC-stable.

On the other extreme, the highest settings of lambda result in very sparse initial graphs which have good precision but poor recall. In general, we see that adding the MGM step increases precision of the PC-stable procedure, at a small cost to recall, depending on the sparsity parameter setting. We see a similar trend in the LD dataset as well (data not shown). In addition, all of our algorithms have both lower precision and recall on edges involving discrete variables which suggests that they are more difficult to learn. These observations differ from the LD setting where we actually achieve the best recall on these *dd* edges, although still diminished precision compared to *cc* and *cd*. Finally, these results show that CPSS is a good option for users that want to ensure very high precision in their network estimates, and is certainly preferable to using an overly sparse setting of lambda.

*Direction recovery.* Next, we evaluated how well each algorithm was able to recover the directions in the directed edges of the true Markov equivalence class. For these tests, the positive class is all estimated directed edges, and the negative class is both undirected edges and the absence of an edge. So, an estimated edge is only considered a true positive if it correctly identifies both the existence and the orientation of the edge. **Figure 2A** shows these results across all of our algorithms. Starting from an MGM graph increases direction recovery performance in PC-stable. The main reason for this improvement appears to be the fact that PC-stable alone returns a large number of bidirected edges and only finds a small number of edges with a single direction. Bidirected edges are returned when the v-structure orientation rule in step 2 of PC-stable implies both directions for an edge. We treat these as undirected edges in our statistics. Starting from an MGM graph reduces the number of bidirected edges and increases the number of directed edge predictions. This is evident by the large increase in directed edge recall, but comes at the price of reduced precision for higher independence test thresholds, $\alpha \in \{.05, .1\}$.

**Figure 2B** gives us a detailed view of the direction recovery performance of CPC-stable, MGM-CPCS, and CPSS. As with adjacency recovery we see that as we increase lambda we achieve higher precision at the cost of recall. The reduced recall in $\lambda \in \{.28, .4\}$ is only slight combined with a significant increase in precision. We can also see that although our heuristic for adapting CPSS to directed network recovery, it is perhaps too conservative as the recall is greatly reduced while precision is near perfect. Indeed, with this set up CPSS predicts the directions of less than 10 edges on average, for the most lenient error rate, $q = .05$, so it does not seem to be a useful option for edge direction predictions.

Overall, direction recovery is difficult in high dimensions. While MGM-PCS approaches direction recall of .3, this is paired with abysmal precision of less than .5. CPC-stable and MGM-CPCS give us reasonable precision, but are able to recall less than 15% of true directed edges. The Matthews correlation We use a strict heuristic to adapt CPSS to the problem of direction estimation that produces extremely high precision.

*Combined measures of network recovery.* The Structural Hamming Distance (SHD) is a combined measure of adjacency and direction, that gives us an alternative network estimation metric that does not necessitate balancing precision versus recall. **Table 1** shows the "best case" performance of our algorithms, where the parameters settings are chosen to maximize the SHD both averaged over all edges, and broken down by each edge type. Since SHD is a distance measure, smaller values indicate better performance. By this measure, MGM-PCS and MGM-CPCS both significantly outperform their counterparts on the HD data. We see a similar trend in the LD data (data not shown), where MGM-PCS performs significantly better than PC-stable, while MGM-CPCS has a slight but non-significant advantage over CPC-Stable.

Since the best case performance will be difficult to achieve when the true graph is unknown, especially in this setting where a robust parameter setting scheme is not readily available, we also show SHD performance versus the number of predicted graph edges. These results, presented in **Figure 3** show that for parameter settings for MGM-CPCS that produce similar numbers of edge predictions to CPC-stable, the hybrid algorithm can improve SHD performance. Very sparse settings of $\lambda$ result in networks with a large SHD because so many edges are missing compared to the true graph. These too-sparse settings of the MGM are evident from the number of predicted edges, however, so they should be easy for a user to identify.

*Run time.* We compared the running times of our algorithms at different parameter settings. **Figure 4** shows these results for the HD dataset. MGM-PCS and MGM-CPCS are significantly faster than PC-stable for sparser settings of $\lambda$, but significantly slower for low values of $\alpha$ and low values of $\lambda$. In the LD data, we see the increase in speed from the MGM step at almost all settings of $\alpha$ and $\lambda$. It is important to note that our MGM learning method is not parallelized, but the directed learning steps are, so a parallelized

MGM learning algorithm could result in even larger speed improvements. The edge convergence approach we use to learning the MGM is essential to this performance improvement.

**TABLES**

**Table 1.** Parameter settings with the best SHD performance by edge type in high-dimensional data set

| Algorithm | α | λ | Type | SHD |
|---|---|---|---|---|
| PC-Stable | 0.01 | none | all | 600.95 (2.25) |
| | 0.01 | none | cc | 130.00 (2.340) |
| | 0.01 | none | cd | 308.40 (4.20) |
| | 0.001 | none | dd | 160.45 (3.24) |
| MGM-PCS | 0.01 | 0.14 | all | 567.75 (3.34) |
| | 0.05 | 0.14 | cc | 108.45 (2.21) |
| | 0.01 | 0.14 | cd | 294.70 (3.74) |
| | 0.001 | 0.1 | dd | 157.30 (3.28) |
| CPC-Stable | 0.01 | none | all | 588.10 (2.37) |
| | 0.05 | none | cc | 111.60 (2.44) |
| | 0.01 | none | cd | 307.05 (4.18) |
| | 0.01 | none | dd | 160.80 (2.85) |
| MGM-CPCS | 0.1 | 0.4 | all | 564.90 (4.46) |
| | 0.1 | 0.57 | cc | 107.05 (2.32) |
| | 0.1 | 0.4 | cd | 296.70 (4.17) |
| | 0.1 | 0.4 | dd | 157.05 (3.25) |

**FIGURE LEGENDS**

**Figure 1.** Precision-Recall curves of edge adjacency recovery on high-dimensional dataset for $0.2 \leq \lambda \leq 0.8$ (represented by different shaped points) and $0.01 \leq \alpha \leq 0.1$. For a given setting of $\lambda$, the different settings of $\alpha$ are connected by lines with colors corresponding to the algorithm. The CPSS line shows the settings of error rate $q \in \{.001, .01, .05, .1\}$.

**Figure 2.** Precision-Recall curves of edge direction recovery on high-dimensional dataset for $0.2 \leq \lambda \leq 0.8$ and $0.01 \leq \alpha \leq 0.1$. **(A)** Full range of algorithms and edge types. **(B)** Detail view of CPC-Stable and MGM-CPC-Stable performance averaged over all edge types.

**Figure 3.** Structural Hamming Distance on high dimensional dataset for CPC-stable and MGM-CPCS with $0.2 \leq \lambda \leq 0.8$ and $0.01 \leq \alpha \leq 0.1$. The lower the SHD, the closer the predicted graph is to the true graph.

**Figure 4.** Average running times with 95% confidence interval error bars of search algorithms on high dimensional data. Each row of bars corresponds to a different setting of $\alpha$ and each color corresponds to a different setting of $\lambda$. Directed search steps were run in parallel on a 4 core laptop.

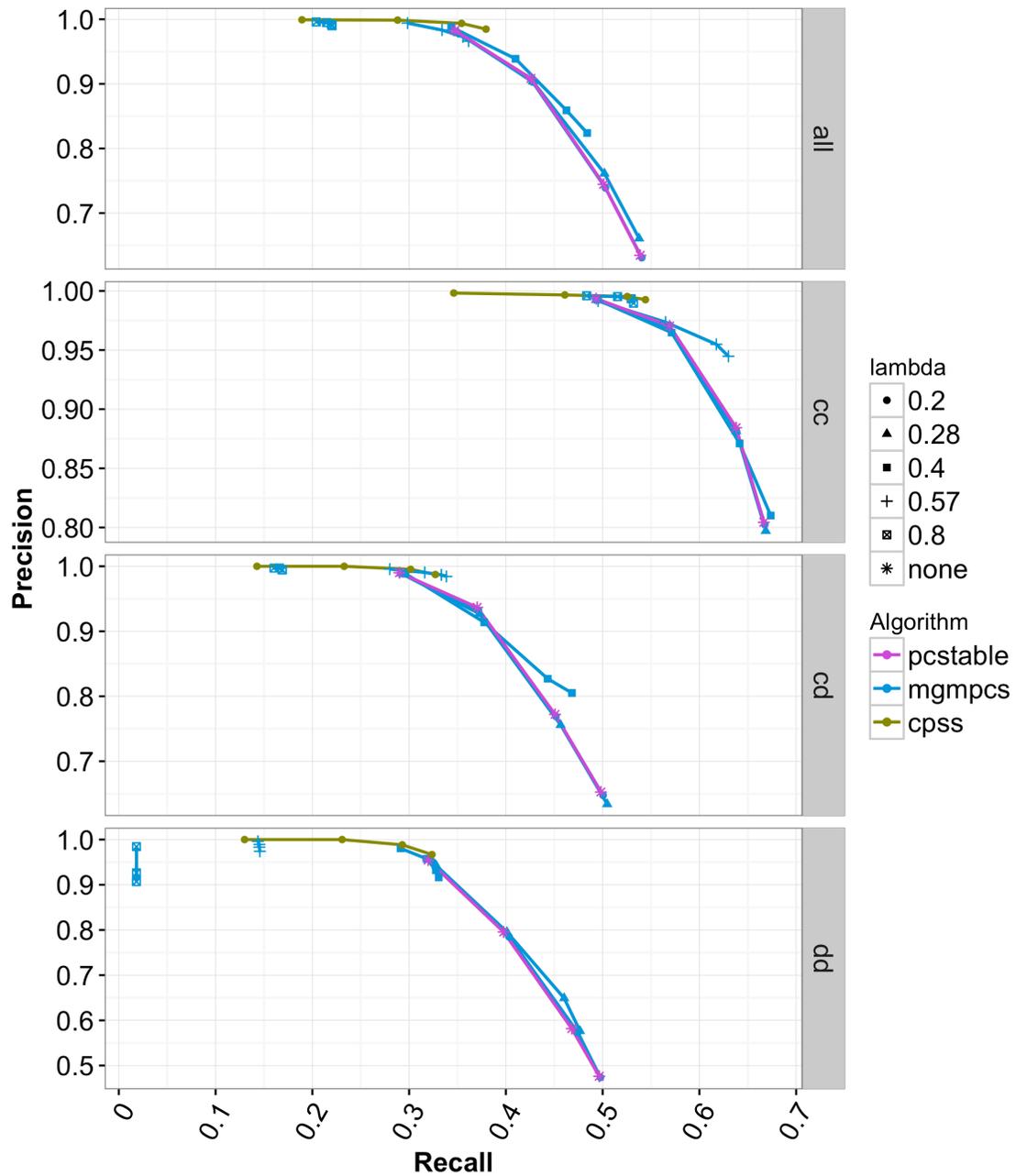

**Figure 1.**

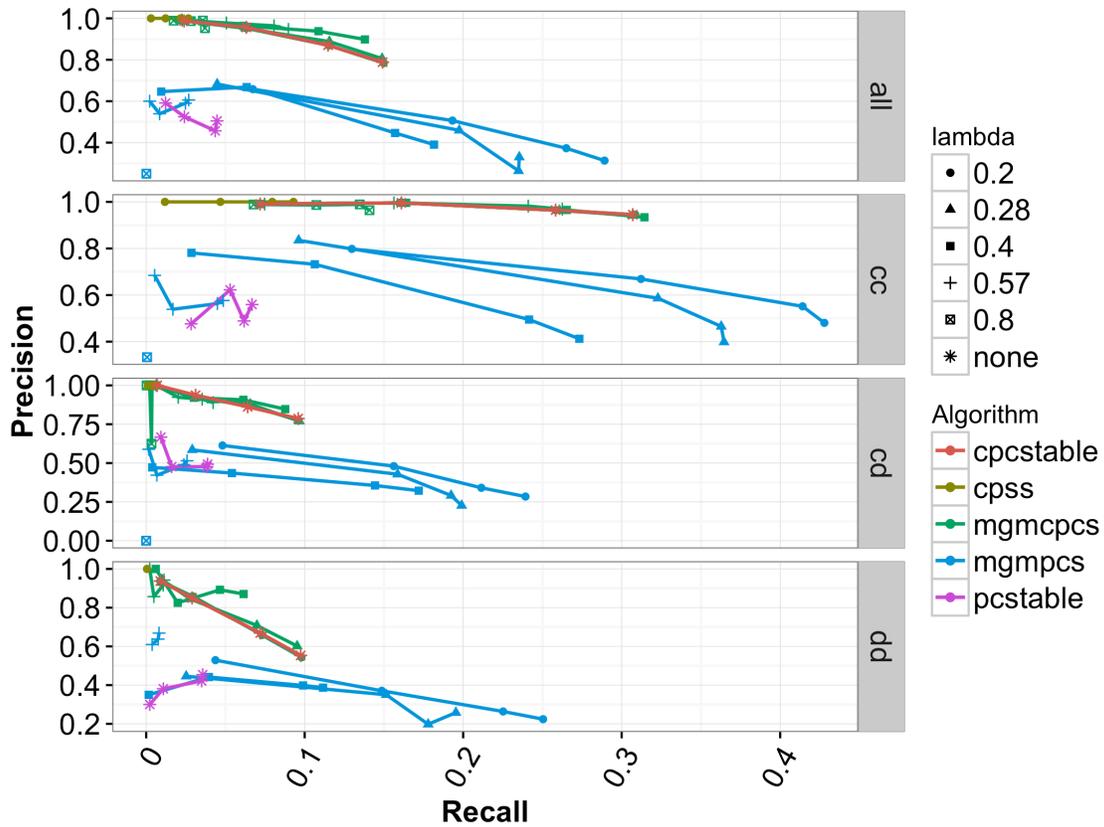

**Figure 2A**

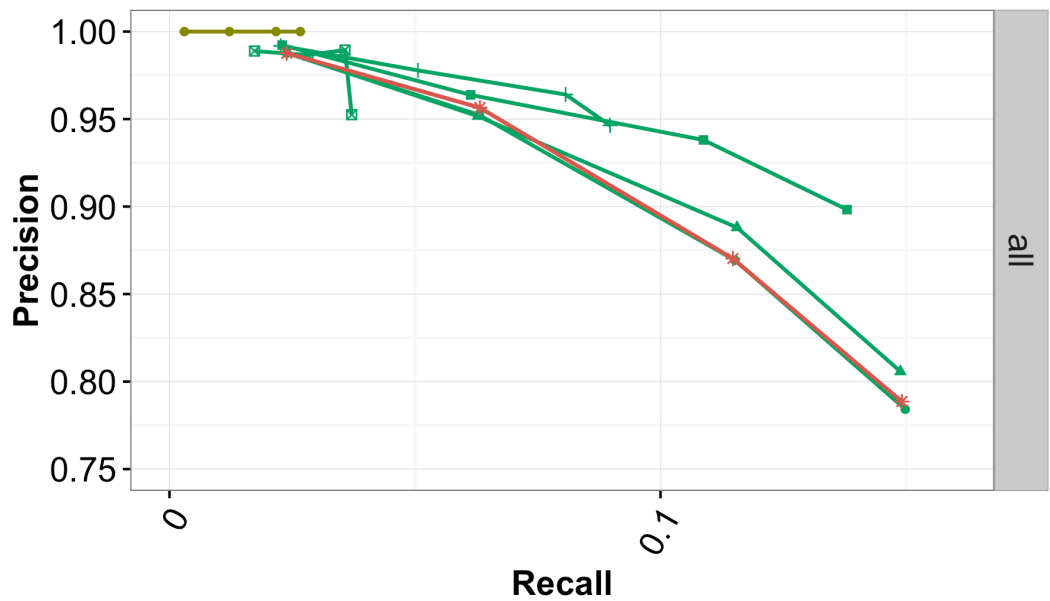

**Figure 2B**

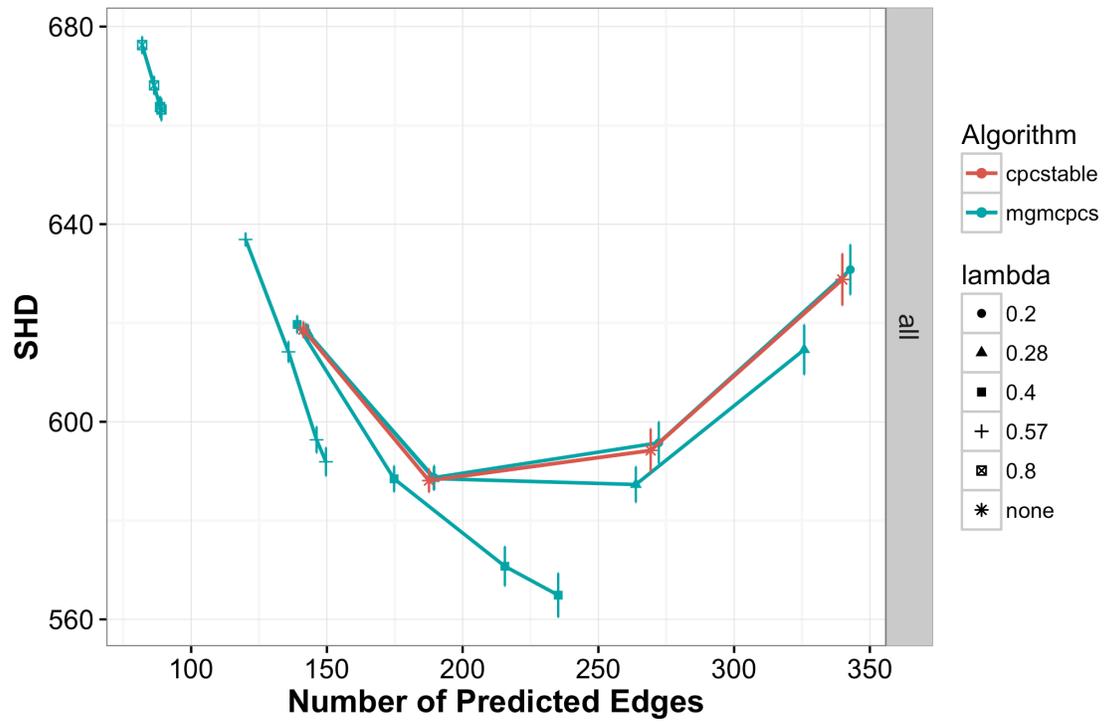

**Figure 3**

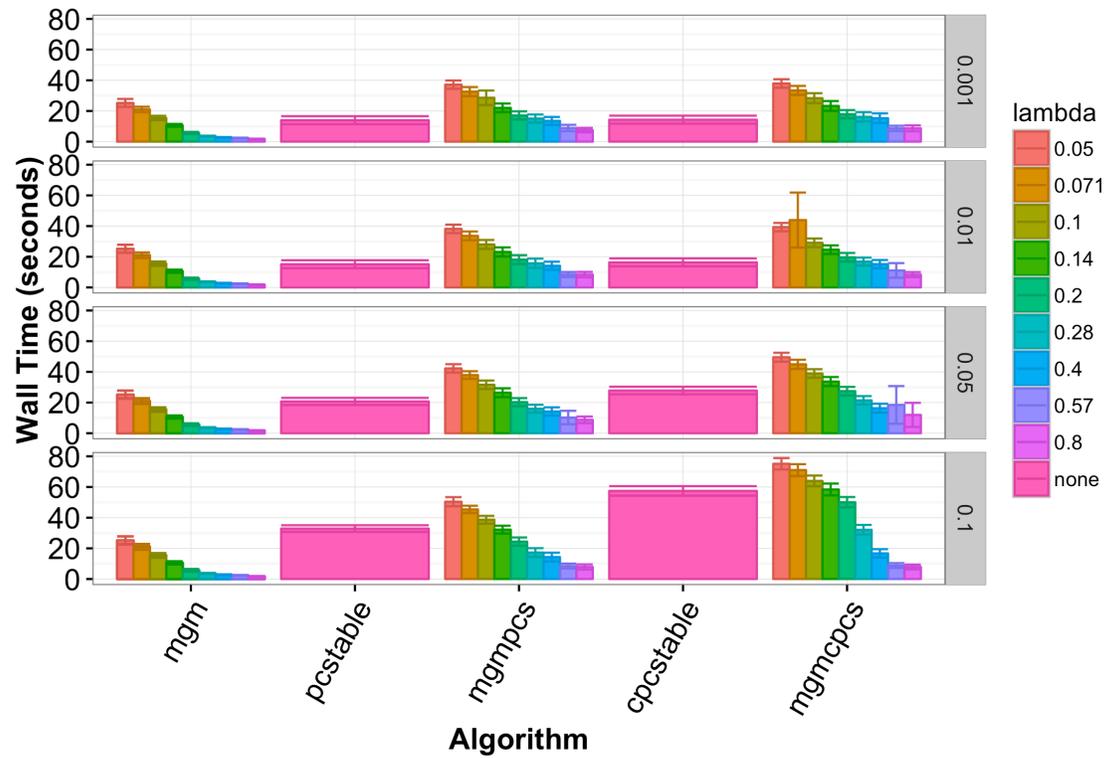

**Figure 4**